\documentclass[sigconf]{acmart}
\usepackage{multirow}
\usepackage{array}
\newcolumntype{C}[1]{>{\centering\arraybackslash}m{#1}}
\AtBeginDocument{%
  }
    
\copyrightyear{2026}
\acmYear{2026}
\setcopyright{cc}
\setcctype{by}
\acmConference[ICMR '26]{International Conference on Multimedia Retrieval}{June 16--19, 2026}{Amsterdam, Netherlands}
\acmBooktitle{International Conference on Multimedia Retrieval (ICMR '26), June 16--19, 2026, Amsterdam, Netherlands}
\acmDOI{10.1145/3805622.3810838}
\acmISBN{979-8-4007-2617-0/2026/06}

\begin{document}

\title{RFHNet: Relational and Frequency-Aware Hashing Network for Large-Scale Fine-Grained Food Image Retrieval}

\author{Junsong Wang}
\affiliation{%
  \institution{College of Computer Science and Artificial Intelligence, Ludong University}
  \city{Yantai}
  \country{China}}
\email{wangjunsong@m.ldu.edu.cn}

\author{Weiqing Min}
\affiliation{%
  \institution{Institute of Computing Technology, Chinese Academy of Sciences}
  \city{Beijing}
  \country{China}}
\email{minweiqing@ict.ac.cn}

\author{Guorui Sheng}
\affiliation{%
  \institution{College of Computer Science and Artificial Intelligence, Ludong University}
  \city{Yantai}
  \country{China}}
\email{shengguorui@ldu.edu.cn}

\author{Tao Yao}
\affiliation{%
  \institution{College of Computer Science and Artificial Intelligence, Ludong University}
  \city{Yantai}
  \country{China}}
\email{yaotao@ldu.edu.cn}

\author{Lili Wang}
\authornote{Corresponding author}
\affiliation{%
  \institution{College of Computer Science and Artificial Intelligence, Ludong University}
  \city{Yantai}
  \country{China}}
\email{wanglili@ldu.edu.cn}

\author{Shuqiang Jiang}
\affiliation{%
  \institution{University of Chinese Academy of Sciences}
  \city{Beijing}
  \country{China}}
\email{sqjiang@ict.ac.cn}

\begin{abstract}
Fine-grained food image retrieval is a key task in computational gastronomy, with applications in food traceability, dietary monitoring, and smart catering systems. Although hashing-based retrieval is attractive for large-scale search due to its storage efficiency and fast Hamming-distance computation, existing methods often perform poorly in fine-grained food scenarios, where subtle local semantics and frequency-sensitive visual cues are essential. To address this challenge, we propose RFHNet, a cascaded hierarchical hashing network that captures both global structure and fine-grained local details through multi-level representations. RFHNet includes three components: (1) Fine-grained Relation Modeling (FRM) to capture subtle visual differences among similar food components; (2) Multi-Frequency Modulated Fusion (MFMF) to extract informative multi-frequency features; and (3) Hierarchical Semantic Synergy (HSS) to adaptively integrate multi-level representations and generate discriminative hash codes. Experiments on six food-specific benchmarks show that RFHNet consistently outperforms state-of-the-art hashing methods, with mAP gains of 4.44\% to 17.20\% at 12 bits. These results validate the effectiveness of RFHNet for large-scale visual food retrieval and smart catering applications. The source code will be released upon publication.
\end{abstract}

\ccsdesc[500]{Information systems~Image search}
\ccsdesc[500]{Computing methodologies~Visual content-based indexing and retrieval}

\keywords{Fine-grained image retrieval, hash learning, multi-frequency information, food computing.}

\maketitle

\section{Introduction}
Food image retrieval is a fundamental task in food computing, supporting applications such as intelligent menu recommendation, dietary monitoring, and nutrition management \cite{min2019survey, FARINELLA201623}. Compared with generic image retrieval, food-related scenarios pose unique challenges due to fine-grained visual variations caused by ingredient composition, cooking styles, and presentation conditions. These challenges are further amplified in large-scale settings, where efficient and discriminative retrieval becomes critical.

\begin{figure}[htbp]
    \centering
    \includegraphics[width=0.7\columnwidth]{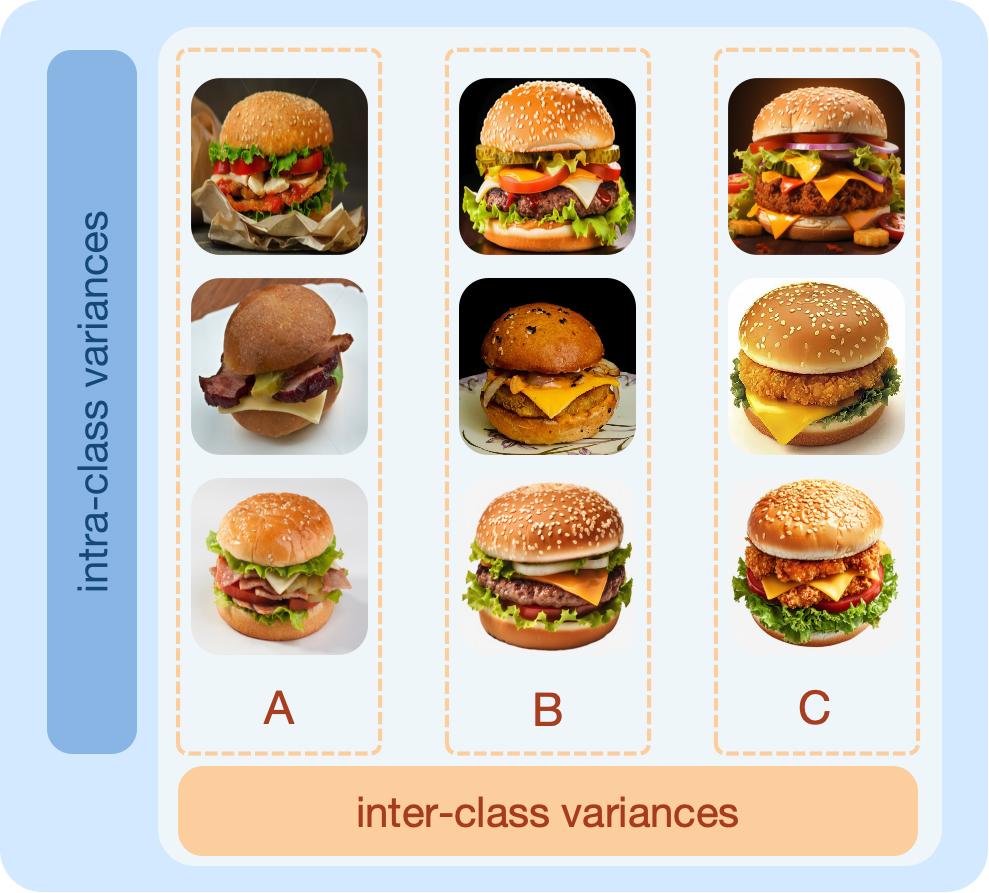}
    \caption{Illustration of inter-class and intra-class variations in fine-grained food image retrieval. Rows show inter-class differences among visually similar categories (A: Bacon Cheeseburger, B: Beef Cheeseburger, C: Chicken Cheeseburger), while columns depict intra-class variations within the same category.}
    \Description{A grid of food images showing inter-class similarity among bacon, beef, and chicken cheeseburgers, and intra-class appearance variations within each category due to ingredients, cooking style, presentation, and viewpoint.}
    \label{fig11}
\end{figure}

With the increasing scale of food image datasets, Fine-Grained Image Retrieval (FGIR) \cite{yu2024gradient} has attracted growing attention. Unlike coarse-grained retrieval \cite{10200622}, FGIR aims to distinguish visually similar instances within the same subcategory, where large intra-class variance and small inter-class variance coexist \cite{wei2021fine}. This problem is particularly severe for food images \cite{bossard2014food, VireoFood172, kawano14c, hou2017vegfru, min2020isia, min2023large}, as visually similar dishes (e.g., beef versus bacon cheeseburgers) and appearance changes caused by preparation styles or viewpoints (Fig.~\ref{fig11}) make discrimination highly challenging. To meet efficiency requirements in large-scale FGIR, hashing-based methods \cite{10.1145/3532624} have been widely adopted. Recent fine-grained hashing approaches \cite{chen2024characteristics, jiang2024global, xiang2024alleviating, li2025semantic} improve retrieval performance by enhancing feature extraction and multi-level fusion. Representative methods, such as A$^2$-Net \cite{NEURIPS2021_2d3acd3e}, A$^2$-Net$^{++}$ \cite{wei2023attribute}, AGMH \cite{lu2023attributes}, and DAHNet \cite{jiang2024global}, exploit attribute modeling, reconstruction, or multi-branch architectures to generate more discriminative hash codes.

Despite these advances, existing methods still face notable limitations in fine-grained food image retrieval. First, many approaches inadequately model subtle spatial relationships between local regions, limiting their ability to distinguish visually similar food categories. Second, most prior methods rely predominantly on spatial-domain feature modeling and are sensitive to both high-frequency noise and low-frequency artifacts, hindering the extraction of stable representations. Furthermore, existing deep hashing frameworks often resort to direct feature concatenation across different regions or layers, implicitly assuming statistical independence and equal contribution of all features. This rigid strategy overlooks the intrinsic semantic correlations and the imbalance of discriminative power among scales, inevitably leading to sub-optimal and less expressive hash codes.

To address these challenges, we propose RFHNet, a hierarchical cascaded hashing network for large-scale fine-grained food image retrieval. RFHNet progressively extracts and fuses multi-level features to jointly model local fine-grained details and global structural information, balancing discriminative power and semantic consistency. Specifically, the proposed framework consists of three key components: (1) Fine-Grained Relational Modeling (FRM) for capturing subtle spatial distinctions between local regions; (2) Multi-Scale Frequency Modulated Fusion (MFMF) for decomposing features into multiple frequency components and suppressing irrelevant noise; and (3) Hierarchical Semantic Synergy (HSS) for adaptively integrating hierarchical features across network stages.

Extensive experiments on six food-specific benchmarks (\textit{Food-101} \cite{bossard2014food}, \textit{Vireo Food-172} \cite{VireoFood172}, \textit{UEC Food-256} \cite{kawano14c}, \textit{VegFru} \cite{hou2017vegfru}, \textit{ISIA Food-500} \cite{min2020isia}, and \textit{Food2K} \cite{min2023large}) demonstrate the effectiveness and robustness of RFHNet.

The main contributions of this work are summarized as follows:
\begin{itemize}
\item We propose a hierarchical cascaded hashing framework that enhances fine-grained food image representation by progressively fusing multi-level features, balancing local discriminability and global consistency.
\item We introduce FRM and MFMF to capture fine-grained spatial and frequency-based distinctions, and devise HSS to adaptively bridge the semantic gap between hierarchical layers for robust cross-scale fusion.
\item Extensive experiments on six food-specific benchmarks demonstrate that RFHNet significantly outperforms state-of-the-art hashing methods, achieving substantial mAP improvements and showing strong robustness across diverse food scenarios.
\end{itemize}

\begin{figure*}[t]
    \centering
    \includegraphics[width=0.9\textwidth]{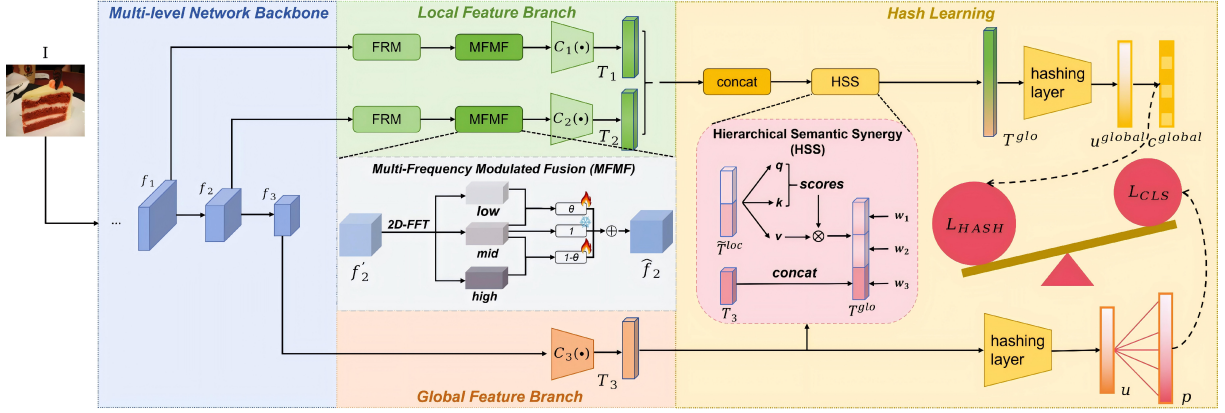}
    \caption{Overall framework of the proposed Relational and Frequency-Aware Hashing Network (RFHNet): Multi-level Backbone, Global Feature Branch, Local Feature Branch, and Hashing Learning.}
    \Description{An overview of the proposed Relational and Frequency-Aware Hashing Network. An input food image is first processed by a multi-level backbone to extract hierarchical features. The global feature branch models overall semantic structure, while the local feature branch captures fine-grained visual details. These representations are then integrated in the hashing learning stage to generate discriminative hash codes for fine-grained food image retrieval.}
    \label{fig1}
\end{figure*}

\section{Related Work}
\subsection{Fine-Grained Image Retrieval (FGIR)}
FGIR aims to distinguish visually similar instances within the same semantic category, and can be broadly divided into fine-grained sketch-based image retrieval (FG-SBIR) \cite{bhunia2021more} and fine-grained content-based image retrieval (FG-CBIR) \cite{dubey2021decade} according to the query modality. FG-SBIR focuses on cross-modal alignment between sketches and images, with an emphasis on local structure and texture modeling. Early methods mainly relied on Siamese or Triplet networks \cite{yu2016sketch, wei2021fine}, whereas recent approaches introduce attention mechanisms and structure-aware representations to improve fine-grained discrimination \cite{sun2022dli, bhunia2022adaptive, chen2022ae}.

FG-CBIR retrieves visually similar images with subtle appearance differences and has been widely studied in product search, species recognition, and food analysis. Traditional methods based on handcrafted features and encoding schemes \cite{lowe2004distinctive, dalal2005histograms, sivic2003video, jegou2010aggregating} are limited in handling complex intra-class variations. Recent deep methods employ multi-branch architectures and attention mechanisms to mine discriminative regions, such as RA-CNN \cite{fu2017look} and MA-CNN \cite{zheng2017learning}, but often incur high computational and storage costs on large-scale datasets.

To improve retrieval efficiency, fine-grained hashing learns compact binary codes while preserving subtle visual distinctions. Representative methods, including ExchNet \cite{cui2020exchnet}, FISH \cite{chen2022fine}, A$^{2}$-Net$^{++}$ \cite{wei2023attribute}, and DAHNet \cite{jiang2024global}, enhance fine-grained hashing through part alignment, filtering, reconstruction, and multi-branch modeling. However, most existing methods generate hash codes by directly concatenating features, overlooking semantic dependencies and the imbalance of feature importance. In contrast, our method adopts attention-guided fusion to model inter-feature relationships before hashing.

\subsection{Hash Learning}
Hash learning accelerates approximate nearest neighbor search by mapping high-dimensional data into compact binary codes. Existing methods are generally divided into data-independent approaches, such as locality-sensitive hashing (LSH) \cite{datar2004locality}, and data-dependent learning-based methods \cite{wang2015learning}. LSH and its variants \cite{lv2007multi, andoni2015optimal} rely on random projections and geometric properties, but often show limited discriminability.

Learning-to-hash methods improve retrieval by optimizing hash functions with supervised or unsupervised data. Early studies mainly focused on distance preservation or label consistency \cite{wang2017survey}, whereas deep hashing methods learn feature representations and hash codes end-to-end, such as DPSH \cite{li2015feature}, HashNet \cite{cao2017hashnet}, and DSH \cite{liu2016deep}. More recent methods introduce attention mechanisms, multi-branch architectures, and transformer-based designs to strengthen semantic representation, including cross-modal hashing \cite{jiang2017deep} and TransHash \cite{chen2022transhash}. Fine-grained hashing is more challenging due to subtle intra-class variations, requiring more expressive feature modeling and fusion strategies, as pursued in this work.

\subsection{Food Image Retrieval}
Food image retrieval is a fundamental task in food computing, supporting efficient visual search for dietary monitoring and nutrition management. Although existing surveys \cite{min2019survey} have extensively reviewed food-related vision tasks, most prior studies focus on recipe retrieval \cite{VireoFood172, zhu2019r2gan, salvador2021revamping}. By comparison, food image retrieval remains relatively underexplored despite its practical value. Early methods mainly relied on metric learning with spatial pyramid features \cite{shimoda2017learning}, whereas more recent approaches combine transformers and convolutional networks to improve robustness and discriminability \cite{song2022noise}. Building on these efforts, we propose a new framework for fine-grained food image retrieval that explicitly models spatial relationships and exploits multi-frequency visual features to enhance discriminability.

\section{Method}

\subsection{Overall Architecture}
We propose RFHNet, a hierarchical cascaded hashing framework for fine-grained food image retrieval, consisting of a multi-stage backbone, a global feature branch, a local feature branch, and a hash learning module (Fig.~\ref{fig1}). A hierarchical feature-extraction backbone based on ResNet-50 is adopted to capture discriminative representations at multiple scales.

Formally, let the backbone be denoted as $\Phi(\cdot;\Theta)$, where $\Theta$ represents the learnable parameters. Given an input image $I$, multi-level feature maps $f=\{f_1,f_2,\ldots,f_J\}$ are extracted from progressively deeper layers:
\begin{equation}
f_j = 
\begin{cases} 
\Phi_1(I, \Theta_1), & j=1, \\
\Phi_j(f_{j-1}, \Theta_j), & j=2,\ldots,J.
\end{cases}
\end{equation}

The deepest feature $f_J$ is directed to the global branch, while the intermediate features $\{f_1,\ldots,f_{J-1}\}$ are processed by the local branch. Within the local branch, for each level $j \in \{1,\ldots,J-1\}$, the FRM and MFMF modules jointly model spatial relationships and multi-frequency characteristics, yielding relational features $f_j^{\prime}$ and frequency-enhanced features $\hat{f}_j$. Finally, convolutional blocks $C_j(\cdot)$ transform these enhanced local features, together with the global feature $f_J$, to generate the final representations $T=\{T_1,\ldots,T_J\}$.

The transformed features are then fused by HSS, which adaptively models cross-scale semantic correlations and aggregates them into a unified representation $T^{\text{glo}}$. The entire network is trained end-to-end with shared parameters to control model complexity. Finally, hash codes are generated via a Tanh activation followed by a sign function, encoding both global structure and fine-grained local details.

\subsection{Fine-Grained Relational Modeling (FRM)}
\label{3b}
\begin{figure}[htbp]
    \centering
    \includegraphics[width=0.65\columnwidth]{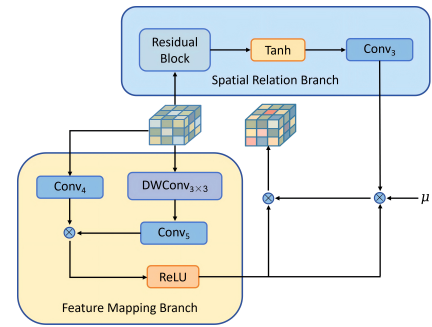}
    \caption{Illustration of Fine-Grained Relational Modeling. The residual block contains two $1 \times 1$ convolutional layers: $\text{Conv}_1$ and $\text{Conv}_2$.}
    \Description{A block diagram of Fine-Grained Relational Modeling. One branch captures local spatial relations through feature projection, subtraction, and nonlinear mapping, while the other preserves local structure through point-wise and depth-wise convolutions. The two branches are fused with a weighted residual connection to generate the refined feature.}
    \label{fig2}
\end{figure}

To capture subtle spatial differences between local regions, we introduce Fine-Grained Relational Modeling (FRM) in the local feature branch (Fig.~\ref{fig2}). FRM consists of a spatial relation branch and a feature mapping branch, jointly modeling inter-region relations and preserving structural information.

Given multi-level features $f_j$, $j\in\{1,\ldots,J-1\}$, the spatial relation branch first reduces the channel dimension by a ratio $r=8$ for efficiency. It then extracts relational features via a difference-based mechanism:
\begin{equation}
f_j^\text{diff}=\text{Conv}_3\!\left(\text{Tanh}(\text{Conv}_1(f_j)-\text{Conv}_2(f_j))\right),
\end{equation}
where $\text{Conv}_1$ and $\text{Conv}_2$ perform the channel reduction, and $\text{Conv}_3$ restores the dimensions. Physically, this subtraction acts as a differential operator, highlighting the contrastive information between different feature projections to capture subtle fine-grained details. Unlike global self-attention mechanisms that incur quadratic computational complexity, this operation efficiently models local relations with minimal overhead.

In parallel, the feature mapping branch preserves local structure using point-wise and depth-wise convolutions:
\begin{equation}
y_1=\text{Conv}_4(f_j), \quad
y_2=\text{Conv}_5(\text{DWConv}_{3\times3}(f_j)),
\end{equation}
where $\text{DWConv}_{3\times3}$ employs padding to maintain spatial alignment. This is followed by non-linear fusion: $y_3=\text{ReLU}(y_1+y_2)$.
Finally, the outputs are fused with a weighted residual formulation:
\begin{equation}
f_j^{\prime}=\mu(f_j^\text{diff}+y_3)+y_3,
\end{equation}
where $\mu$ is a learnable fusion coefficient initialized to 1.0 to adaptively integrate the relational cues.

\subsection{Multi-Frequency Modulated Fusion (MFMF)}
\label{3c}
\begin{figure}[htbp]
    \centering
    \includegraphics[width=\columnwidth]{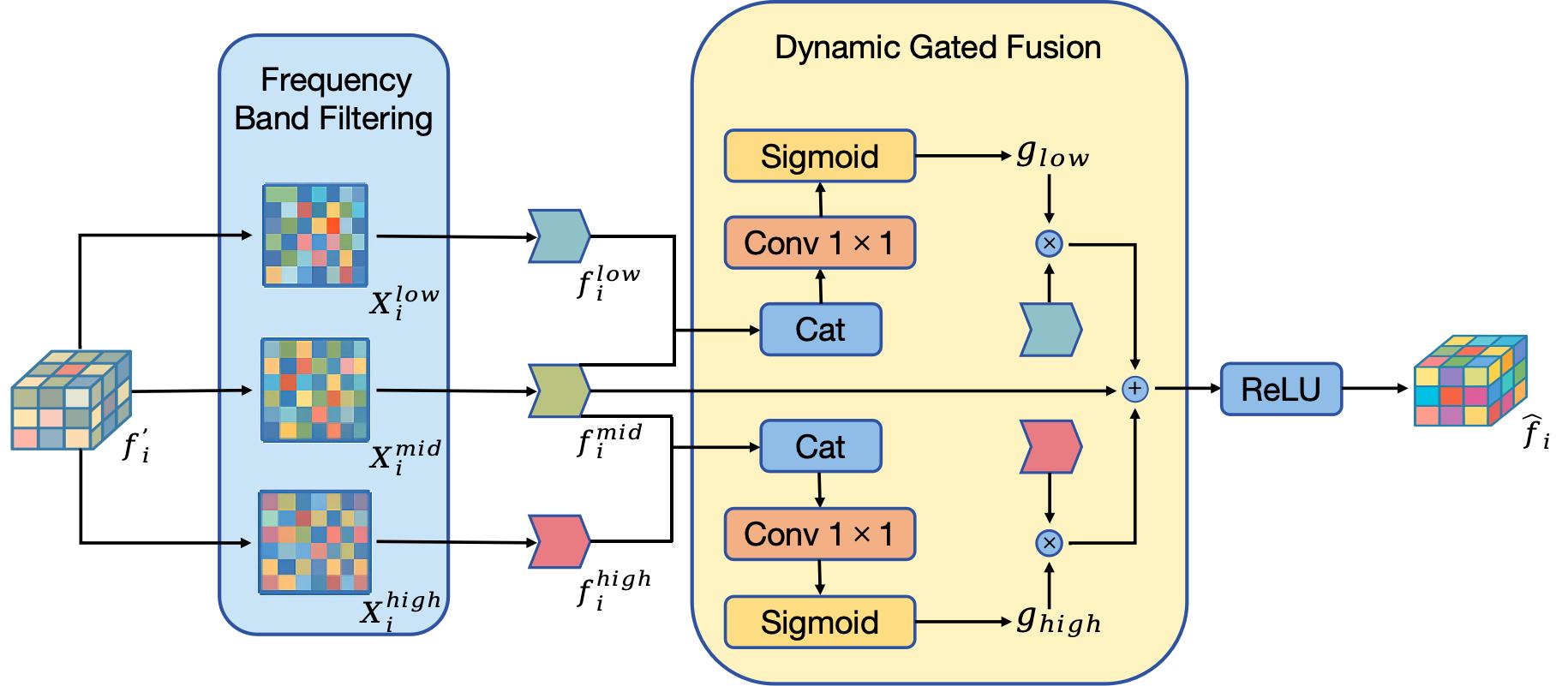}
    \caption{Overview of Multi-Frequency Modulated Fusion.}
    \Description{A block diagram of Multi-Frequency Modulated Fusion. An input feature is decomposed into low-, mid-, and high-frequency components in the Fourier domain, transformed back to the spatial domain, and then adaptively fused. The mid-frequency feature guides gating of the low- and high-frequency components, and the three components are combined to form the final fused feature.}
    \label{fig3}
\end{figure}

To address the limitations of spatial-domain modeling and filter irrelevant noise, we propose the Multi-Frequency Modulated Fusion (MFMF) (Fig.~\ref{fig3}). Leveraging 2D-FFT and a dynamic gating mechanism, the MFMF adaptively fuses frequency components to generate robust food feature representations.

Given an input feature map $f_j^{\prime}\in\mathbb{R}^{B\times C\times S\times S}$, we apply 2D-FFT to obtain its frequency representation $X_j^f=\text{FFT2D}(f_j^{\prime})$. To precisely partition the spectrum, we define $d(u,v)$ as the Euclidean distance from the spectrum center to position $(u,v)$, and $R$ as the maximum radius of the spectrum. By setting a frequency threshold ratio $\tau$ (empirically set to 0.3), the binary masks $M^*$ are generated as follows:
\begin{equation}
M^*_{(u,v)} = \begin{cases} 
1, & \text{if } *=\text{low} \text{ and } d(u,v) \leq \tau R, \\
1, & \text{if } *=\text{mid} \text{ and } \tau R < d(u,v) < (1-\tau) R, \\
1, & \text{if } *=\text{high} \text{ and } d(u,v) \geq (1-\tau) R, \\
0, & \text{otherwise}.
\end{cases}
\end{equation}

Then, the frequency bands are obtained by $X_j^*=X_j^f\odot M^*$, where $*\in\{\text{low, mid, high}\}$. These bands capture global shapes, textures, and fine details (or noise), respectively. Each band is transformed back to the spatial domain:
\begin{equation}
f_j^*=\text{Real}(\text{IFFT2D}(X_j^*))\in\mathbb{R}^{B\times C\times S\times S}.
\end{equation}
To adaptively fuse these components, we employ a gating mechanism using mid-frequency features as the contextual anchor to guide the enhancement:
\begin{equation}
\begin{aligned}
g_\text{low} &= \sigma(\text{Conv}_{1\times1}([f_j^\text{mid}; f_j^\text{low}])), \\
g_\text{high} &= \sigma(\text{Conv}_{1\times1}([f_j^\text{mid}; f_j^\text{high}])),
\end{aligned}
\end{equation}
where $\sigma(\cdot)$ is the Sigmoid function. Finally, the fused feature is obtained by enhancing low- and high-frequency bands while preserving the structural consistency from the mid-frequency band:
\begin{equation}
\widehat{f}_j=g_\text{low}\odot f_j^\text{low}+f_j^\text{mid}+g_\text{high}\odot f_j^\text{high}.
\end{equation}
A ReLU activation is subsequently applied.

\subsection{Hierarchical Semantic Synergy (HSS)}
\label{3d}
To synergize local details with global context while preventing semantic dilution, we introduce the Hierarchical Semantic Synergy (HSS). By decoupling local feature interaction from global integration, HSS explicitly models intrinsic cross-scale correlations (or fine-grained layer-wise dependencies) before unifying them with the holistic representation.

Given the feature set $T=\{T_1,\ldots,T_J\}$, where $T_J$ denotes the global feature and the subset $\{T_1,\ldots,T_{J-1}\}$ represents local features, we first isolate the local branch to model fine-grained correlations. The local features are stacked to form a sequence:

\begin{equation}
T^\text{loc}=\text{Stack}(T_1,\ldots,T_{J-1})\in\mathbb{R}^{B\times (J-1)\times D}.
\end{equation}

To capture region-to-region dependencies, self-attention is performed exclusively on the local sequence $T^\text{loc}$. The query, key, and value projections are computed as:

\begin{equation}
Q=T^\text{loc}W_Q,\quad K=T^\text{loc}W_K,\quad V=T^\text{loc}W_V,
\end{equation}
followed by the scaled dot-product attention to generate the enhanced local features $\widetilde{T}^\text{loc}$:
\begin{equation}
\widetilde{T}^\text{loc}=\text{Softmax}\left(\frac{QK^\top}{\sqrt{D}}\right)V \in\mathbb{R}^{B\times (J-1)\times D}.
\end{equation}

Subsequently, to reintegrate holistic semantic information, the enhanced local features are concatenated with the preserved global feature $T_J$, yielding the complete feature set:

\begin{equation}
\widetilde{T}=\text{Concat}(\widetilde{T}^\text{loc}, T_J)\in\mathbb{R}^{B\times J\times D}.
\end{equation}

Finally, a learnable weight vector $\mathbf{W}\in\mathbb{R}^J$ is applied to adaptively balance the contribution of the refined local details and the global context. The fused representation is obtained by weighting and flattening the features:

\begin{equation}
T^\text{glo}=\text{Reshape}\left(\widetilde{T} \odot \text{Softmax}(\mathbf{W})^\top\right)\in\mathbb{R}^{B\times JD},
\end{equation}
where $\odot$ denotes channel-wise broadcasting multiplication.

\begin{table*}
\centering
\caption{Comparison of mAP (\%) with state-of-the-art methods on six fine-grained food datasets.}
\label{tab:1}

\small 

\begin{tabular}{@{} C{1.8cm} | C{0.8cm} | C{0.9cm} C{0.9cm} C{0.9cm} C{1.1cm} C{1.2cm} C{0.9cm} C{1.1cm} C{1.0cm} C{1.2cm} | C{0.8cm} @{} }
\toprule

Datasets & $\#$ bits & HashNet & ADSH & A$^{2}$-Net & SEMICON & A$^{2}$-Net$^{++}$ & AMGH & DAHNet & SPBH & FoodHash & \textbf{Ours}\\
\midrule

\multirow{4}{*}{\textit{Food-101}} 
& 12 & 24.42 & 35.64 & 46.44 & 50.00 & 54.51 & 62.59 & 67.67 & \underline{80.70} & 77.26 & \textbf{85.14}\\
& 24 & 34.48 & 40.93 & 66.87 & 76.57 & 81.46 & 80.94 & 79.38 & \underline{86.07} & 83.01 & \textbf{86.17}\\
& 32 & 35.90 & 42.89 & 74.27 & 80.19 & 82.91 & 82.31 & 82.05 & \textbf{86.57} & 82.62 & \underline{86.44}\\
& 48 & 39.65 & 48.81 & 82.13 & 82.44 & 83.66 & 83.21 & 83.11 & \textbf{87.53} & 83.44 & \underline{86.42}\\
\midrule

\multirow{4}{*}{\textit{Vireo Food-172}} 
& 12 & 1.65 & 20.80 & 37.95 & 32.03 & 38.25 & 36.66 & 36.73 & 68.13 & \underline{75.55} & \textbf{83.14}\\
& 24 & 2.34 & 38.86 & 64.45 & 59.89 & 68.49 & 68.70 & 60.43 & 78.79 & \underline{84.73} & \textbf{86.44}\\
& 32 & 2.96 & 43.99 & 71.40 & 65.25 & 75.73 & 73.95 & 68.24 & 79.83 & \underline{84.96} & \textbf{86.78}\\
& 48 & 4.22 & 61.15 & 76.68 & 69.87 & 79.42 & 76.56 & 75.39 & 81.22 & \underline{85.11} & \textbf{87.35}\\
\midrule

\multirow{4}{*}{\textit{UEC Food-256}} 
& 12 & 3.23 & 11.58 & 24.65 & 14.40 & 20.00 & 24.97 & 41.75 & \underline{59.09} & 48.52 & \textbf{65.45}\\
& 24 & 10.35 & 19.32 & 47.83 & 35.66 & 41.43 & 50.21 & 63.89 & \textbf{72.05} & 63.88 & \underline{69.23}\\
& 32 & 15.64 & 21.88 & 58.52 & 39.67 & 47.01 & 59.97 & 68.05 & \textbf{71.93} & 64.57 & \underline{70.22}\\
& 48 & 16.33 & 30.48 & 67.60 & 55.99 & 58.40 & 69.57 & \textbf{72.85} & \underline{72.64} & 66.12 & 71.42\\
\midrule

\multirow{4}{*}{\textit{VegFru}} 
& 12 & 3.70 & 8.24 & 25.52 & 30.32 & 30.54 & 43.99 & 56.11 & 63.14 & \underline{65.83} & \textbf{83.03}\\
& 24 & 6.24 & 24.90 & 44.73 & 58.45 & 60.56 & 68.05 & 78.07 & \underline{80.60} & 80.33 & \textbf{86.28}\\
& 32 & 7.83 & 36.53 & 52.75 & 69.92 & 73.38 & 76.73 & 82.19 & \underline{83.02} & 81.83 & \textbf{86.60}\\
& 48 & 10.29 & 55.15 & 69.77 & 79.77 & 82.80 & 84.49 & 85.56 & \underline{84.32} & 82.76 & \textbf{87.14}\\
\midrule

\multirow{4}{*}{\textit{ISIA Food-500}} 
& 12 & 0.77 & 4.93 & 10.09 & 5.36 & 10.29 & 11.29 & 5.93 & 15.40 & \underline{32.03} & \textbf{48.01}\\
& 24 & 4.11 & 10.40 & 17.37 & 7.93 & 17.78 & 20.26 & 17.43 & 27.18 & \underline{47.28} & \textbf{53.85}\\
& 32 & 6.26 & 12.64 & 21.27 & 8.94 & 21.53 & 23.62 & 31.23 & 28.42 & \underline{48.91} & \textbf{54.58}\\
& 48 & 8.08 & 15.87 & 26.61 & 22.16 & 26.41 & 29.02 & 38.12 & 29.47 & \underline{50.09} & \textbf{55.54}\\
\midrule

\multirow{4}{*}{\textit{Food2K}} 
& 12 & 0.18 & 0.21 & 3.94 & 1.25 & 3.94 & 5.42 & 0.14 & 1.72 & 26.73 & \textbf{43.66}\\
& 24 & 0.83 & 1.03 & 9.05 & 1.96 & 9.22 & 8.64 & 0.19 & 7.70 & \textbf{67.01} & \underline{63.45}\\
& 32 & 3.17 & 2.50 & 12.00 & 2.76 & 12.22 & 9.22 & 0.36 & 11.21 & \textbf{67.03} & \underline{65.15}\\
& 48 & 4.42 & 4.85 & 15.95 & 3.50 & 15.66 & 12.64 & 1.94 & 15.32 & \textbf{71.15} & \underline{66.47}\\
\bottomrule
\end{tabular}
\end{table*}
\subsection{Loss Function}
As shown in Fig.~\ref{fig1}, RFHNet is trained using a multi-task objective that jointly optimizes classification and hashing, enabling collaborative learning of semantic discrimination and compact hash codes.

\subsubsection{Classification Loss}
We adopt the cross-entropy loss with label smoothing to improve generalization. Let $\mathbf{y}_i \in \{0,1\}^K$ denote the one-hot ground-truth vector for the $i$-th image, where $K$ represents the number of classes. The smoothed label distribution $\tilde{y}_{i,k}$ for the $k$-th class is defined as:
\begin{equation}
\tilde{y}_{i,k} = (1-\lambda)y_{i,k} + \frac{\lambda}{K},
\end{equation}
where $\lambda$ is the smoothing parameter (set to 0.1) and $y_{i,k}$ is the $k$-th element of $\mathbf{y}_i$ (1 for the target class, 0 otherwise). Consequently, the classification loss is formulated as:
\begin{equation}
L_\text{CLS} = - \sum_{i=1}^N \sum_{k=1}^K \tilde{y}_{i,k} \log(p_{i,k}),
\end{equation}
where $p_{i,k}$ is the predicted probability.

\subsubsection{Hash Loss}
\label{3e2}
Following FISH \cite{chen2022fine}, we adopt a proxy-based hashing loss for stable optimization in fine-grained hashing. Let $c_i=H^{\prime}(T_i^\text{glo})$ denote the hash code of image $i$, and $d_u$ be the proxy of class $u$. The relaxed proxy objective is given by:
\begin{equation}
\begin{aligned}
L_\text{hash}^{\prime} = &\ \|Y-DC\|_F^2 + \sum_{i=1}^n \|c_i - W^h T_i^\text{glo}\|_F^2, \\
&\text{s.t.}\quad C \in \{-1,1\}^{k \times n}.
\end{aligned}
\end{equation}
where $D$ denotes the class prototypes, $C$ is the binary hash code matrix, $W^h$ is the projection weight of the hashing layer, and $c_i$ is the target code. The optimization is decoupled into two steps:
\begin{equation}
L_\text{HASH}=\sum_{i=1}^n\|c_i-W^hT_i^\text{glo}\|_F^2.
\end{equation}

The offline hash code learning is formulated as:
\begin{equation}
\min_{C,D}\|Y-DC\|_F^2,\quad \text{s.t.}\quad C\in\{-1,1\}^{k\times n},
\end{equation}
which is solved via relaxation and orthogonal rotation following FISH.

\subsubsection{Total Loss and Hash Code Generation}
To balance classification and hashing tasks automatically without manual tuning, we employ a multi-task loss with learnable uncertainties:
\begin{equation}
\begin{aligned}
L_\text{TOTAL}=&\frac{1}{\alpha^2}L_\text{HASH}+\frac{1}{\beta^2}L_\text{CLS} \\
&+\log(\alpha+1)+\log(\beta+1),
\end{aligned}
\end{equation}
where $\alpha$ and $\beta$ are learnable parameters optimized jointly with the network.

After training, the final binary hash code for a query image is obtained as:
\begin{equation}
C_q=\text{sign}(W^hT_q^\text{glo}),
\end{equation}
which compactly encodes semantic and fine-grained information for efficient retrieval.

\begin{table*}
\centering
\caption{Food retrieval accuracy (\% mAP) with incremental components of the proposed RFHNet model.}
\label{tab:3}
\begin{tabular}{@{} C{0.16\textwidth} | C{0.08\textwidth} | C{0.18\textwidth} C{0.12\textwidth} C{0.12\textwidth} C{0.12\textwidth} @{}}
\toprule
\multirow{2}{*}{Datasets} & \multirow{2}{*}{$\#$ bits} & Vanilla Backbone & +FRM & +MFMF & +HSS \\
& & (ResNet-50) & (Sec. 3.2) & (Sec. 3.3) & (Sec. 3.4) \\
\midrule
\multirow{4}{*}{\textit{Food-101}} & 12 & 83.30 & 84.12 & 85.03 & 85.14 \\
& 24 & 85.11 & 85.31 & 85.89 & 86.17 \\
& 32 & 86.15 & 86.21 & 86.38 & 86.44 \\
& 48 & 85.54 & 85.77 & 86.30 & 86.42 \\
\midrule
\multirow{4}{*}{\textit{Vireo Food-172}} & 12 & 80.48 & 81.22 & 82.73 & 83.14 \\
& 24 & 84.66 & 85.83 & 86.28 & 86.44 \\
& 32 & 84.59 & 85.23 & 86.45 & 86.78 \\
& 48 & 85.39 & 86.25 & 87.23 & 87.35 \\
\midrule
\multirow{4}{*}{\textit{UEC Food-256}} & 12 & 62.77 & 63.17 & 64.53 & 65.45 \\
& 24 & 67.55 & 68.13 & 68.62 & 69.23 \\
& 32 & 67.84 & 68.03 & 69.83 & 70.22 \\
& 48 & 68.63 & 70.02 & 70.85 & 71.42 \\
\midrule
\multirow{4}{*}{\textit{VegFru}} & 12 & 81.33 & 82.04 & 82.86 & 83.03 \\
& 24 & 84.61 & 85.11 & 86.03 & 86.28 \\
& 32 & 85.01 & 85.67 & 86.25 & 86.60 \\
& 48 & 86.69 & 86.60 & 86.97 & 87.14 \\
\midrule
\multirow{4}{*}{\textit{ISIA Food-500}} & 12 & 42.70 & 44.05 & 46.72 & 48.01 \\
& 24 & 51.17 & 51.99 & 53.32 & 53.85 \\
& 32 & 52.31 & 53.08 & 53.94 & 54.58 \\
& 48 & 52.56 & 53.93 & 55.49 & 55.54 \\
\midrule
\multirow{4}{*}{\textit{Food2K}} & 12 & 38.58 & 40.36 & 43.17 & 43.66 \\
& 24 & 61.33 & 62.38 & 63.29 & 63.45 \\
& 32 & 63.77 & 63.91 & 64.82 & 65.15 \\
& 48 & 64.02 & 64.98 & 66.27 & 66.47 \\
\bottomrule
\end{tabular}
\end{table*}
\section{Experiments}

\subsection{Datasets}
We evaluate RFHNet on six large-scale benchmarks, strictly adhering to official protocols: \textit{Food-101} \cite{bossard2014food} (101k images, 101 classes, standard 750/250 split); \textit{Vireo Food-172} \cite{VireoFood172} (110k images, 172 classes, random 80\%/20\% split); \textit{UEC Food-256} \cite{kawano14c} ($\sim$32k images, 256 classes, using valid bounding boxes); \textit{VegFru} \cite{hou2017vegfru} ($>$160k images, 292 categories, official train/val/test split); \textit{ISIA Food-500} \cite{min2020isia} (399k images, 500 classes, fixed 800/200 split); and \textit{Food2K} \cite{min2023large} ($>$1M images, 2,000 categories, standard split).

\subsection{Evaluation Metric}
We evaluate retrieval performance using mean Average Precision (mAP). Specifically, we report mAP@1000, which computes the Average Precision (AP) over the top 1,000 retrieved results for each query and averages the AP scores across all queries. This metric is widely used to assess effectiveness in large-scale retrieval scenarios.

\begin{table*}
\centering
\caption{Comparison of mAP (\%) between proposed RFHNet (based on Resnet18 and Resnet50) and DAHNet (based on Resnet50).}
\label{tab:4} 
\begin{tabular}{@{} C{0.18\textwidth} | C{0.08\textwidth} | C{0.16\textwidth} C{0.22\textwidth} C{0.22\textwidth} @{}}
\toprule
Datasets & $\#$ bits & DAHNet & RFHNet$_{\text{ResNet18}}$ & RFHNet$_{\text{ResNet50}}$\\
\midrule
\multirow{4}{*}{\textit{Food-101}} & 12 & 67.67 & 80.02 & 85.14\\
& 24 & 79.38 & 82.72 & 86.17\\
& 32 & 82.05 & 82.76 & 86.44\\
& 48 & 83.11 & 83.22 & 86.42\\
\midrule
\multirow{4}{*}{\textit{Vireo Food-172}} & 12 & 36.73 & 78.25 & 83.14\\
& 24 & 60.43 & 84.50 & 86.44\\
& 32 & 68.24 & 84.62 & 86.78\\
& 48 & 75.39 & 85.19 & 87.35\\
\midrule
\multirow{4}{*}{\textit{UEC Food-256}} & 12 & 41.75 & 56.79 & 65.45\\
& 24 & 63.89 & 64.19 & 69.23\\
& 32 & 68.05 & 65.86 & 70.22\\
& 48 & 72.85 & 67.67 & 71.42\\
\midrule
\multirow{4}{*}{\textit{VegFru}} & 12 & 56.11 & 76.97 & 83.03 \\
& 24 & 78.07 & 81.58 & 86.28 \\
& 32 & 82.19 & 81.97 & 86.60 \\
& 48 & 85.56 & 82.98 & 87.14 \\
\midrule
\multirow{4}{*}{\textit{ISIA Food-500}} & 12 & 5.93 & 39.18 & 48.01 \\
& 24 & 17.43 & 46.59 & 53.85 \\
& 32 & 31.23 & 48.13 & 54.58 \\
& 48 & 38.12 & 49.68 & 55.54 \\
\midrule
\multirow{4}{*}{\textit{Food2K}} & 12 & 0.14 & 20.13 & 43.66 \\
& 24 & 0.19 & 28.77 & 63.45 \\
& 32 & 0.36 & 31.49 & 65.15 \\
& 48 & 1.94 & 33.08 & 66.47 \\
\bottomrule
\end{tabular}
\end{table*}

\begin{table*}
\centering
\caption{mAP (\%) on the testing set with fixed hyperparameters of loss functions on \textit{UEC Food-256}.}
\label{tab:hpy} 
\begin{tabular}{ @{} C{0.15\textwidth} | C{0.10\textwidth} C{0.10\textwidth} | C{0.12\textwidth} C{0.12\textwidth} C{0.12\textwidth} C{0.12\textwidth} @{} }
\toprule
\multirow{2}{*}{Method} & \multicolumn{2}{c|}{Parameter} & \multicolumn{4}{c}{\textit{UEC Food-256}} \\
& $\alpha$ & $\beta$ & 12bits & 24bits & 32bits & 48bits \\
\midrule
\multirow{5}{*}{Fixed} & 0.8 & 1 & 62.08 & 66.46 & 67.24 & 69.42 \\
& 1 & 0.8 & 62.67 & 67.57 & 67.09 & 69.92 \\
& 1 & 1 & 62.94 & 67.40 & 68.67 & 70.14 \\
& 0.5 & 1 & 63.00 & 68.54 & 69.27 & 70.89 \\
& 1 & 0.5 & 62.37 & 68.67 & 68.93 & 71.11 \\
\midrule
Dynamic & - & - & \textbf{65.45} & \textbf{69.23} & \textbf{70.22} & \textbf{71.42} \\
\bottomrule
\end{tabular}
\end{table*}
\begin{figure*}[htbp]
    \centering
    \includegraphics[width=0.65\textwidth]{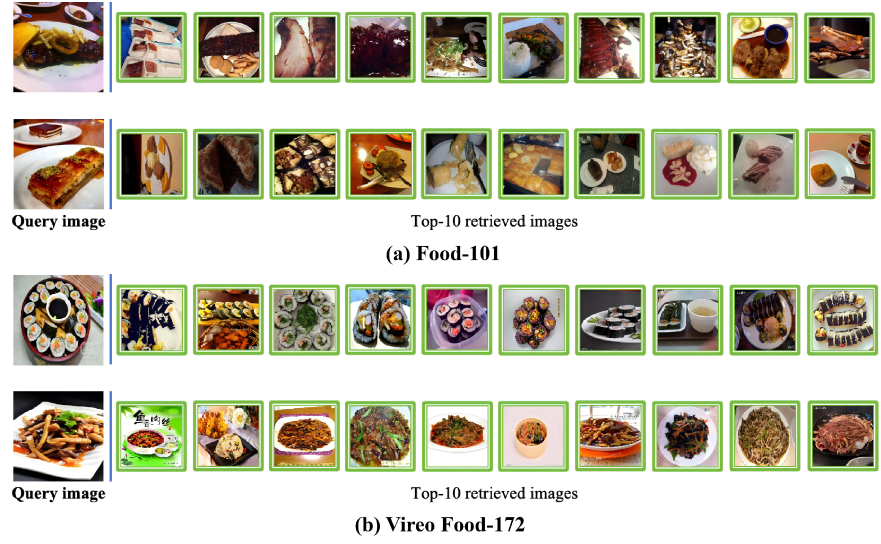}
    \caption{Examples of the top-10 retrieved images using 48-bit hash codes generated by RFHNet on two representative fine-grained food image datasets (\textit{Food-101} and \textit{Vireo Food-172}).}
    \Description{Examples of image retrieval results produced by RFHNet using 48-bit hash codes on the Food-101 and Vireo Food-172 datasets. Each query food image is followed by its top-10 retrieved images. The figure illustrates that the retrieved results are visually and semantically similar to the query, showing the effectiveness of RFHNet on fine-grained food image retrieval across two representative datasets.}
    \label{re}
\end{figure*}

\subsection{Comparison Methods and Experimental Settings}

\subsubsection{Comparison Methods}
We compare RFHNet with nine representative hashing methods, including HashNet \cite{he2016deep} and ADSH \cite{jiang2018asymmetric} for coarse-grained retrieval, and seven state-of-the-art fine-grained hashing methods: A$^2$-Net \cite{NEURIPS2021_2d3acd3e}, SEMICON \cite{shen2022semicon}, A$^2$-Net$^{++}$ \cite{wei2023attribute}, AMGH \cite{lu2023attributes}, DAHNet \cite{jiang2024global}, SPBH \cite{li2025semantic}, and FoodHash \cite{cao2026foodhash}.

\begin{figure*}[htbp]
    \centering
    \includegraphics[width=0.7\textwidth]{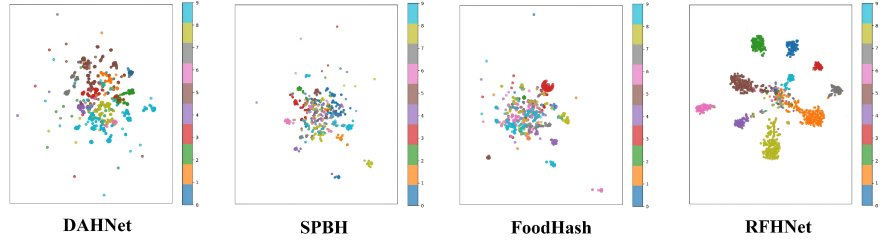}
    \caption{t-SNE visualization of 12-bit hash codes learned by DAHNet, SPBH, FoodHash, and the proposed RFHNet on the \textit{VegFru} dataset.}
    \Description{A t-SNE visualization of 12-bit hash codes produced by DAHNet, SPBH, FoodHash, and RFHNet on the VegFru dataset. Each subplot shows the distribution of hash representations in two-dimensional space, with different colors indicating different classes. The figure is used to compare class compactness and class separation among the four methods.}
    \label{t}
\end{figure*}

\begin{table}[htbp]
\centering
\caption{Inference efficiency comparison between RFHNet and FoodHash on four large-scale fine-grained food datasets. GPU Memory (MB) and Retrieval Time (ms/img) denote peak memory usage and average latency per image, respectively. $\downarrow$ indicates lower is better.}
\label{tab:eff}
\footnotesize 
\setlength{\tabcolsep}{3pt} 
\begin{tabular}{@{} l c c c c @{}}
\toprule

\multirow{2}{*}[-0.5ex]{\textbf{Dataset}} & \multicolumn{2}{c}{\textbf{GPU Memory (MB)} $\downarrow$} & \multicolumn{2}{c}{\textbf{Retrieval Time (ms/img)} $\downarrow$} \\
\cmidrule(lr){2-3} \cmidrule(lr){4-5} 
 & FoodHash & \textbf{RFHNet (Ours)} & FoodHash & \textbf{RFHNet (Ours)} \\
\midrule
\textit{Food-101}       & 11550.5 & \textbf{1651.5} & 0.726 & \textbf{0.282} \\
\textit{Vireo Food-172} & 11183.3 & \textbf{1659.9} & 0.731 & \textbf{0.283} \\
\textit{UEC Food-256}   & 11536.6 & \textbf{1648.0} & 0.732 & \textbf{0.286} \\
\textit{Food2K}         & 25527.5 & \textbf{11855.5} & 0.748 & \textbf{0.305} \\
\bottomrule
\end{tabular}
\end{table}

\subsubsection{Experimental Settings}
For fair comparison, all CNN-based baselines adopt ResNet-50 or ResNet-18 backbones pre-trained on ImageNet. For the Transformer-based method FoodHash, we employ ViT-Small to ensure comparable model complexity. All backbones have their original pooling and classification layers removed, and only image-level category labels are used for supervision. During training, images are randomly cropped and resized to $224\times224$ with horizontal flipping; during testing, images are resized to $256\times256$ and center-cropped to $224\times224$. Models are trained using SGD with momentum 0.9 and weight decay $3\times10^{-4}$. Unless otherwise specified, the learning rate is 0.008, batch size is 128, and training lasts 300 epochs. For \textit{ISIA Food-500}, training is reduced to 200 epochs; for \textit{VegFru}, the learning rate is 0.005; and for \textit{Food2K}, training is conducted for 150 epochs.

\subsection{Performance Comparison}
Table~\ref{tab:1} compares RFHNet with state-of-the-art methods. RFHNet establishes a new benchmark for short-code hashing, achieving the best performance on \textit{all six} datasets at 12 bits. Compared to the strongest competitors (SPBH and FoodHash), RFHNet yields substantial improvements in the 12-bit setting: 17.20\% on \textit{VegFru}, 16.93\% on \textit{Food2K}, and 15.98\% on \textit{ISIA Food-500}. Significant leads are also achieved on \textit{Vireo Food-172} (+7.59\%), \textit{UEC Food-256} (+6.36\%), and \textit{Food-101} (+4.44\%). Moreover, on datasets like \textit{Vireo Food-172}, \textit{VegFru}, and \textit{ISIA Food-500}, our method consistently outperforms benchmarks across all bit lengths (12-48 bits), validating its effectiveness in capturing discriminative fine-grained features.

\subsection{Ablation Study}
\subsubsection{Ablation on Model Components}
We progressively integrate FRM, MFMF, and HSS into a vanilla cascaded ResNet-50 backbone and evaluate their effects on six food datasets. As shown in Table~\ref{tab:3}, retrieval performance consistently improves with each module added, demonstrating the effectiveness and complementarity of all components.

\subsubsection{Ablation on Backbone}
To evaluate performance under lightweight settings, we replace ResNet-50 with ResNet-18. As reported in Table~\ref{tab:4}, RFHNet with ResNet-18 consistently outperforms DAHNet (ResNet-50) across all datasets, while stronger backbones further improve performance, indicating good scalability.

\subsubsection{Ablation on Loss Function}
We evaluate the impact of learnable loss weights on \textit{UEC Food-256}. As shown in Table~\ref{tab:hpy}, models with learnable parameters consistently outperform those with fixed settings, confirming the benefit of adaptive loss balancing.

\subsection{Qualitative Results}
Fig.~\ref{re} presents representative retrieval results, demonstrating robust performance under diverse backgrounds and fine-grained variations. Fig.~\ref{t} visualizes the t-SNE embeddings of 12-bit hash codes on \textit{VegFru}. Compared with DAHNet, SPBH, and FoodHash, the proposed RFHNet exhibits tighter intra-class clustering and larger inter-class margins. This qualitative result confirms that our method effectively captures subtle visual semantics, generating more discriminative representations for fine-grained retrieval.
\subsection{Inference Efficiency Analysis}
\label{Efficiency}
Inference efficiency is crucial for large-scale deployment. As shown in Table~\ref{tab:eff}, RFHNet requires fewer resources than the Transformer-based FoodHash while preserving real-time retrieval performance. This benefit mainly stems from the Hierarchical Semantic Synergy (HSS) module, which avoids the quadratic complexity of Vision Transformers through lightweight feature interaction.

\section{Conclusion}
We proposed RFHNet, a hierarchical cascaded hashing network for large-scale fine-grained food image retrieval. By jointly modeling fine-grained spatial relations and multi-frequency characteristics, RFHNet learns compact and discriminative hash codes that preserve both local details and global structure. Extensive experiments on six fine-grained food benchmarks demonstrate consistent performance gains, particularly excelling in 12-bit short-code retrieval scenarios.

\bibliography{ref}

\end{document}